\documentclass[runningheads]{llncs}
\usepackage{graphicx}
\usepackage{subfigure}
\usepackage{tabularx}
\usepackage{hyperref}
\usepackage{pdflscape}
\usepackage{caption}
\usepackage{amsmath} 
\begin{document}
\title{Agricultural Robotic System: The Automation of  Detection and Speech Control\thanks{This research project is partially supported by the Ministry of Education, Singapore, under its Research Centre of Excellence award to the Institute for Functional Intelligent Materials (I-FIM, project No. EDUNC-33-18-279-V12). The real robot presentation and discription will show in this link. \url{https://www.youtube.com/watch?v=S4Op68Es7FY}}}
%
%
\author{Yang Wenkai\inst{1}
Ji Ruihang
Yue Yiran
Gu Zhonghan 
Shu Wanyang \and
Sam Ge Shuzhi}

%
%
\institute{National University of Singapore, Singapore 
\email{e0983030@u.nus.edu}}

\maketitle 
%
%

%
\begin{abstract}
Agriculture industries often face challenges in manual tasks such as planting, harvesting, fertilizing, and detection, which can be time-consuming and prone to errors. The "Agricultural Robotic System" project addresses these issues through a modular design that integrates advanced visual, speech recognition, and robotic technologies. This system is comprised of separate but interconnected modules for vision detection and speech recognition, creating a flexible and adaptable solution. The vision detection module uses computer vision techniques, trained on YOLOv5 and deployed on the Jetson Nano in TensorRT format, to accurately detect and identify different items. A robotic arm module then precisely controls the picking up of seedlings or seeds, and arranges them in specific locations. The speech recognition module enhances intelligent human-robot interaction, allowing for efficient and intuitive control of the system. This modular approach improves the efficiency and accuracy of agricultural tasks, demonstrating the potential of robotics in the agricultural industry.

\end{abstract}

\section{Introduction}
This paper discusses the challenges in agriculture and how robotics and computer vision technologies can be used to address them. Specifically, the impact of the Agricultural Robotic System (ARS) on the efficiency of agricultural processes is addressed. We in this work focus on addressing the ARS with object detection, speech recognition and robot arm trajectory. The ARS is a system that uses advanced computer vision techniques yolov5 \cite{yan2021real} to accurately detect, identify, and differentiate between various plant species and their growth stages. The system is designed to be highly efficient and adaptable, and can be scaled to accommodate different farm sizes and types of crops.

The ARS can be used to automate a variety of agricultural tasks, including planting, harvesting, fertilizing, and detection. For example, the ARS can be used to plant seeds with precision and accuracy, ensuring that the seeds are planted in the correct location and at the correct depth. The ARS can also be used to harvest crops with minimal damage, and can even be used to identify and remove diseased or damaged plants. The ARS can also be used to apply fertilizer to crops in a precise and efficient manner, ensuring that the fertilizer is applied in the correct amount and in the correct location.

The ARS has the potential to significantly improve the efficiency and productivity of agricultural processes. The system can automate many of the labor-intensive tasks that are currently performed by humans, and can do so with greater precision and accuracy. The ARS can also help to reduce the environmental impact of agriculture, by reducing the use of pesticides and herbicides.

The ARS is still under development, but it has the potential to revolutionize the agricultural industry. The system has the potential to make farming more efficient, productive, and sustainable. One of the key challenges in deploying the ARS is optimizing the model for the device. This requires striking a balance between model accuracy and inference speed. Too much optimization may lead to loss of accuracy, while too little may not significantly improve inference speed. Another challenge is implementing Non-Maximum Suppression (NMS) on the Nano. NMS is a computationally expensive process and can slow down the overall inference time.

\section{Analysis of Agricultural Robotic system}
\begin{figure}
  \begin{center}
  \includegraphics[width=\textwidth]{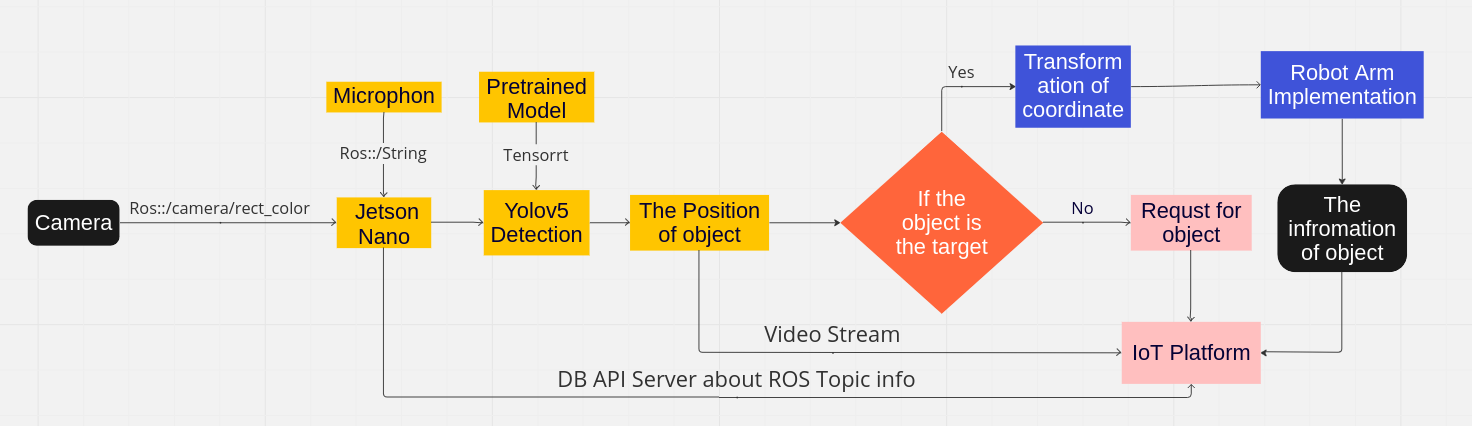}\\
  \caption{The Agricultural Robotic System System Design}
  \label{System_architecture}
  \end{center}
\end{figure}
Consider the system architecture shown in Fig.~\ref{System_architecture}.  The RGB camera captures high-quality images and transmits them in the ROS (Robot Operating System) format, enabling the integration of visual information into our complex robotic ecosystem. Concurrently, microphone system is designed to pick up and analyze human speech. For instance, a command such as "pick the orange" is processed and translated into a series of actions for the robotic arm. Meanwhile, the NVIDIA Jetson Nano is a powerful device that loads a pre-trained machine learning model in TensorRT \cite{jocher2022ultralytics} and processes video streams in real time.The raw result from the initial inference is refined using non-maximum suppression (NMS), which reduces overlapping bounding boxes and narrows down the focus to the most probable object locations. The result of NMS is transformed into the operational space of the robotic arm. This allows the arm to calculate and follow an optimal path to reach and pick up the detected object.

\section{Pattern Recognition and Robot Intelligence}

\subsection{Computer Vison Detection}

We use Yolov5 to detect fruits and seeds in the video from the robotic arm. Yolov5 is a fast, accurate, and lightweight object detection algorithm developed by Ultralytics. It is suitable for deployment on resource-constrained devices such as mobile phones and robots. Yolov5 is a one-stage object detection algorithm that uses a single CNN to perform object detection. It consists of a backbone network, neck network, and head network. The backbone network extracts features from the input image, the neck network enhances the discriminability of the features, and the head network predicts the bounding boxes and class probabilities of the objects in the image. The backbone network is based on the CSP Darknet53 \cite{wang2021scaled} architecture, which has been optimized for efficiency and accuracy. The CSP block allows for efficient information flow across multiple layers. The detection head in YOLOv5 is a hybrid approach, which combines anchor-based \cite{zhang2020bridging} and anchor-free methods to improve the detection performance. It uses a SPP \cite{he2015spatial} (spatial pyramid pooling) module to extract features at multiple scales, followed by a PAN \cite{liu2018path} (path aggregation network) module that combines features from different scales. Finally, a YOLOv5 head predicts objectness scores, bounding box coordinates, and class probabilities for each anchor box.
Yolov5 is an anchor-based object\cite{tian2020fcos} detection algorithm that uses a multi-scale\cite{ren2008multi} approach to detect objects of different sizes. The network predicts the bounding box coordinates, class probabilities, and offset of each anchor box from the true object location at different scales of the feature map.

The dataset used for training the Yolov5 model contains 200 images, each of which has one or more of the following four types of objects: apple, banana, orange, and seed. The dataset is balanced, with approximately equal numbers of each object type. Specifically, there are 202 apples, 181 bananas, 178 oranges, and 146 seeds in the dataset. To get a better understanding of the dataset, we can also look at the distribution of objects across the images. As mentioned, each image can contain one or more objects. The following Table \ref{Dataset} summarizes the number of objects per image:
\begin{table}
\caption{Objects Distribution}
\label{Dataset}
\centering
\begin{tabular}{|c|c|}
\hline
Number of objects & Number of images \\
\hline
1 & 24 \\
2 & 46 \\
3 & 36 \\
4 & 80 \\
5 & 39 \\
6 & 8 \\
\hline
\end{tabular}
\end{table}
From the Table \ref{Dataset}, we can see that the majority of the images (60 out of 200) contain 4 objects, while the fewest images (8 out of 200) contain 6 objects. During the training process, Yolov5 model uses data augmentation techniques such as random flipping and cropping to increase the size of the dataset and improve the model's ability to generalize to new images.

\subsection{Speech Recognition}

Speech recognition is a process of pattern recognition. It works by splitting the waveform into utterances by silences, then trying to recognize what's being said in each utterance. The best matching combination is chosen. Speech recognition modules based on acoustic models are designed to achieve more convenient human-computer interaction.
The speech recognition module used a pre-trained Chinese acoustic model from CMUSphinx \cite{lamere2003cmu}. The model has been optimized for optimal performance and can be used in most instruction interaction systems, even for applications with large vocabularies. Sphinx and some open-source pre-trained models can be used to adapt to existing models. We installed the SpeedRecognition and PocketSphinx \cite{huggins2006pocketsphinx} modules, which achieved high accuracy and fast speed. We can consider training our own model for higher accuracy, but the existing model has already met the initial requirements

A continuous speech recognition system can be roughly divided into four parts: feature extraction, acoustic model training, language model training, and decoding.
\begin{itemize}
    \item Feature extraction: MFCC features \cite{murty2005combining} are used in Sphinx. MFCC is a widely used parameter due to its excellent noise resistance and robustness. It is calculated using FFT \cite{frigo1998fftw}, convolution, and DCT.
    \item Acoustic model training: HMM \cite{eddy2011accelerated} is a statistical model established for the time series structure of speech signals. We use a left to right unidirectional, self loop, and spanning topology to model recognition primitives. A phoneme is an HMM in three to five states, a word is an HMM composed of multiple phonemes that form a word in series, and the entire model of continuous speech recognition is an HMM composed of words and silence.
    \item Language model training: In a traditional stochastic language model, the current word is predicted based on the preceding word or the preceding n-l words. We experimented with long distance bigrams to reduce the number of free parameters and maintain the modeling capacity.
    \item Decoder: The recognition result of a continuous speech recognition system is a word sequence. Decoding is actually a repeated search for all the words in the vocabulary. The arrangement of words in a vocabulary can affect the speed of search.
\end{itemize}

We have set up the vocabulary to be loaded into the module, and then we can start recording and recognizing the vocabulary input by the user, and return a list of possible matching Chinese vocabulary. The controller makes corresponding actions and operations based on these vocabulary and the previously established vocabulary-action mapping.


\subsection{Image Geometry}

After we get the bounding box of every object in pixel coordinate, we need to transform it into world coordinate. This is a common task in computer vision and involves a series of transformations that map the points from the image plane (pixels) to the real-world 3D coordinates. The procedure typically involves the following steps:

\subsubsection{Camera Calibration}

Camera calibration is the process of estimating the intrinsic and extrinsic parameters of a camera. The intrinsic parameters are related to the camera's internal characteristics, such as focal length, optical center, and lens distortion. The extrinsic parameters are the camera's position and orientation in the world. This step is essential because it establishes a relationship between pixel coordinates and world coordinates.

\subsubsection{Pixel Coordinates to Camera Coordinates}

The first step in conversion is to map the pixel coordinates to the camera's coordinate system. The equation for this is given by:
\begin{equation}
\begin{bmatrix}
u \\
v \\
1
\end{bmatrix}
=
\begin{bmatrix}
f_x & 0 & c_x \\
0 & f_y & c_y \\
0 & 0 & 1
\end{bmatrix}
\begin{bmatrix}
X_c \\
Y_c \\
Z_c
\end{bmatrix}
\end{equation}
where $(u, v)$ are the pixel coordinates, $(X_c, Y_c, Z_c)$ are the coordinates in the camera frame, $(fx, fy)$ are the focal lengths in $x$ and $y$ directions (in pixel units), and $(cx, cy)$ is the optical center (in pixel units). This is actually a pinhole camera model which is a good approximation for many real cameras.

\subsubsection{Camera Coordinates to World Coordinates}

Once we have the point in the camera coordinate system, we can transform it to the world coordinate system using the extrinsic parameters. The equation is given by:
\begin{equation}
\begin{bmatrix}
X_w \\
Y_w \\
Z_w
\end{bmatrix}
=
\begin{bmatrix}
R & t \\
0 & 1
\end{bmatrix}
\begin{bmatrix}
X_c \\
Y_c \\
Z_c
\end{bmatrix}
\end{equation}
where ($X_w, Y_w, Z_w)$ are the world coordinates, $R$ is a 3x3 rotation matrix, and $t$ is a 3x1 translation vector. The `0` in the bottom left is a row vector of size 1x3 and the `1` in the bottom right is a scalar. This is a homogeneous transformation matrix. These values are obtained from the camera calibration process and define the camera's position and orientation in the world.

The above process require the camera have been already calibrated and know its intrinsic and extrinsic parameters. Hence we perform camera calibration first using a method the chessboard calibration technique \cite{de2010automatic}.

\subsection{Robot Arm Kinematics and Dynamics}

Robot arm kinematics \cite{lee1982robot} is the study of the motion of robotic arms. It is used to determine the position and orientation of the end-effector for a given set of joint angles, and to compute the joint angles required to achieve a desired position and orientation of the end-effector. The robot kinematics can be divided into forward kinematics and inverse kinematics. The forward kinematics is to calculate the position of the end of the robot given the joint angles of the manipulator. Inverse kinematics means that the position of the robot end is known and all joint angles of the corresponding position of the manipulator are calculated. In practical application, the inverse kinematics is used to calculate the joint Angle given the position of the end of the manipulator.
In general, forward kinematics has a unique solution and is easy to obtain, while inverse kinematics often has multiple solutions and is more complex to analyze. The region where the inverse kinematics solutions exist is usually referred to as the robot's workspace. The workspace is the collection of target points that the robot arm's end-effector can reach.

\subsubsection{Inverse kinematics calculation}

\begin{figure}
  \begin{center}
  \includegraphics[width=3.4in]{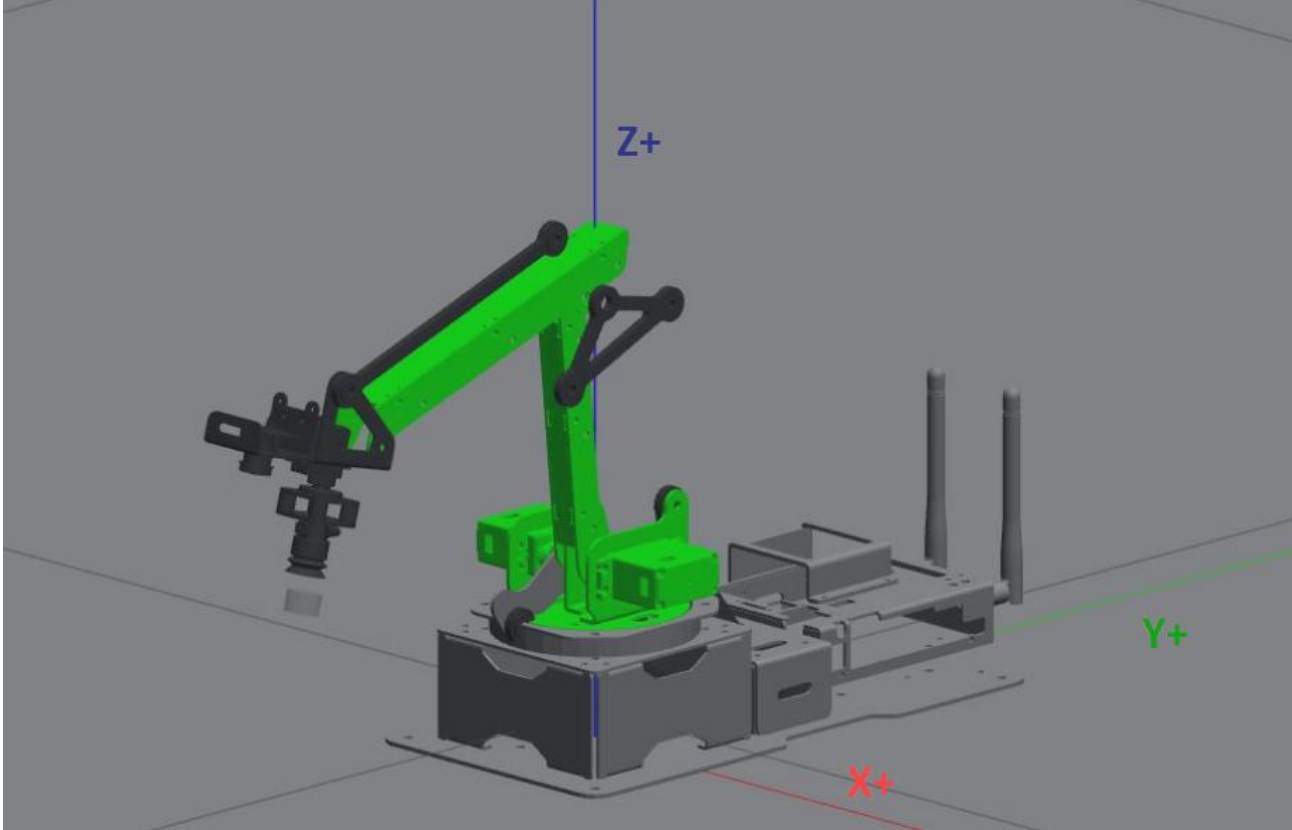}\\
  \caption{Coordinate System of Robot Arm}
  \label{Coordinate}
  \end{center}
\end{figure}

Let's assume the joint angles are represented by $\theta_1$, $\theta_2$, and $\theta_3$, and the link lengths are $L_1$, $L_2$, and $L_3$. The desired position of the end-effector is given by the coordinates $(x,y,z)$. Secondly, we set up the cartesian coordinate system in Figure \ref{Coordinate} to describe spatial position, velocity, and acceleration. Our JetMax arm uses X, Y and Z. The three-axis coordinate system takes the center point of the mechanical arm base as the origin of the coordinate system $(0,0,0)$, as shown in Figure \ref{Coordinate}. When the manipulator moves to the right, the value of $x$ decreases. When the manipulator moves forward, the value of $y$ decreases. As the manipulator moves up, the value of $z$ increases.

First, let's find the position of the end-effector center (point $W$) in the coordinate system. The distance from the wrist center to the end-effector is $L_3$. We can write the coordinates of the wrist center as:$L_3$. We can write the coordinates of the wrist center as:
\begin{equation}
\begin{aligned}
    W_x &= x - L_3 \cos(\theta_3), \\
    W_y &= y - L_3 \sin(\theta_3), \\
    W_z &= z.
\end{aligned}
\end{equation}
Next, we can find the first joint angle $\theta_1$ as the angle between the x-axis and the projection of the wrist center onto the xy-plane:$\theta_1$as the angle between the x-axis and the projection of the $\theta_1$ as
\begin{equation}
    \theta_1 = \mathrm{atan2}(W_y, W_x)
\end{equation}
Now, we can find the distance from the first joint to the wrist center:
\begin{equation}
    D = \sqrt{W_x^2 + W_y^2 + (W_z - L_1)^2}
\end{equation}
Using the Law of Cosines, we can find the second and third joint angles:
\begin{equation}
\begin{aligned}
    & \cos(\theta_2) = \frac{L_2^2 + D^2 - L_3^2}{2 L_2 D} \\
    & \sin(\theta_2) = \sqrt{1 - \cos^2(\theta_2)} \\
    & \theta_2 = \mathrm{atan2}(\sin(\theta_2), \cos(\theta_2))
\end{aligned}
\end{equation}
\begin{equation}
\begin{aligned}
    & \cos(\theta_3) = \frac{L_2^2 + L_3^2 - D^2}{2 L_2 L_3} \\
    & \sin(\theta_3) = \sqrt{1 - \cos^2(\theta_3)} \\
    & \theta_3 = \mathrm{atan2}(\sin(\theta_3), \cos(\theta_3))
\end{aligned}
\end{equation}
According to these equations provide the inverse kinematics solution for a 3-DoF robotic manipulator, we can directly set the position from jetson nano in python code.

\section{The Practical Experiments of Robot}

\subsection{Computer Vision Detection}

\begin{figure}[ht!]
\centering
\includegraphics[width=3.4in]{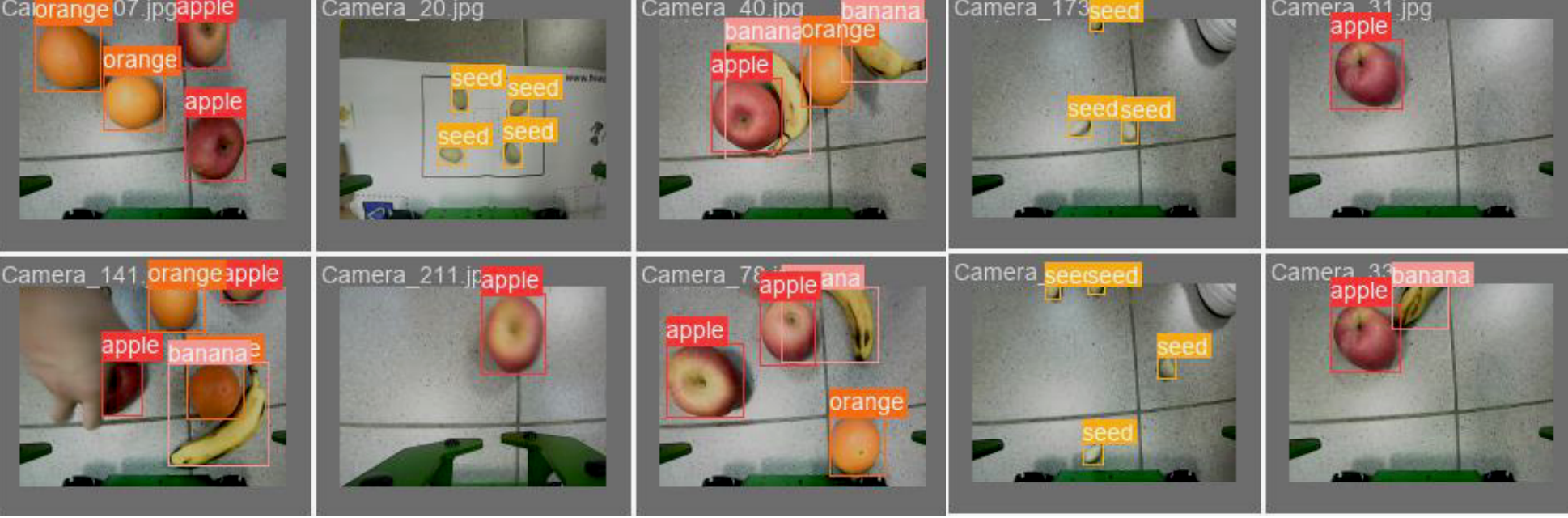}
\caption{Validation of Yolov5}
\label{val}
\end{figure}

\begin{figure}[ht!]
\centering
\includegraphics[width=3.5in]{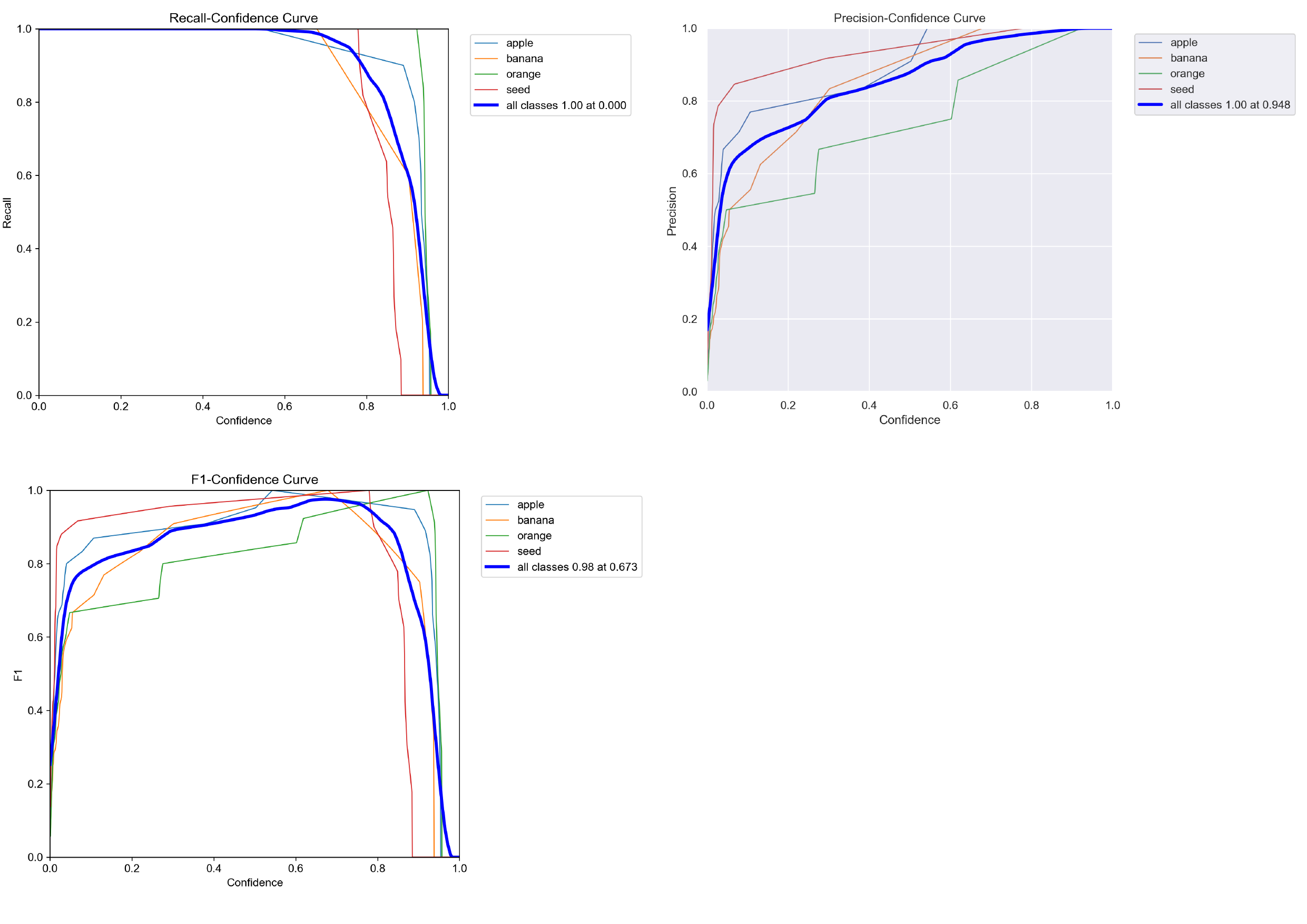}
\caption{Recall, Precision and F1 of Yolov5}
\label{R_P_F}
\end{figure}

The Yolov5 model achieved an average f1 score of 0.98 for all 4 classes (apple, banana, orange, and seed) on the validation showed in Figure \ref{val} and \ref{R_P_F} set when the confidence threshold was set to 0.67. This is a high level of accuracy, indicating that the model is able to accurately detect and classify the objects in the images. A confidence threshold of 0.67 means that the model is only considering detections that it is relatively confident in, with a score of at least 0.67 out of 1. 


\subsection{Evaluation of Agricultural Robotic System}
The robot arm has demonstrated significant capabilities in the realm of agricultural technology. It can pick up seedlings or seeds with precise control, harvest crops, and identify items using advanced computer vision techniques. The speech recognition module and IoT interface enhance its user-friendliness and adaptability. Future enhancements could include refining the object detection model, expanding the acoustic models, and exploring more complex movements and operations. The Table \ref{tab:robot_arms} provides a concise comparison of various robot arms, highlighting their item picking abilities and special features. Overall, the Agricultural Robotic project stands as a beacon of innovation in agricultural technology.
\begin{table}
\caption{Comparison of Robot Arms}
\label{tab:robot_arms}
\begin{tabularx}{\linewidth}{|c|X|X|}
\hline
\textbf{Robot Arm} & \textbf{Item Picking Ability} & \textbf{Special Features} \\ \hline
Agricultural Robotic & Picks up different seedlings or seeds and harvests crops. & Uses computer vision and speech recognition. The IoT interface display ROS topic information in real time API server. \\ \hline
ABB IRB 360 FlexPicker & Known for speed and precision in picking and packing. & Uses ABB's PickMaster software with vision guidance. \\ \hline
Fanuc M-2iA/3S & Designed for high-speed picking and packing. & Uses Fanuc's iRVision system. \\ \hline
Universal Robots UR5 & Equipped with a camera for vision-guided tasks. & Known for flexibility and safety features. \\ \hline
KUKA KR AGILUS & Equipped with vision systems for picking tasks. & Known for speed and precision. \\ \hline
Yaskawa Motoman HC10 & Equipped with a camera for vision-guided picking. & Designed to work safely alongside humans. \\ \hline
\end{tabularx}
\end{table}
\section{Conclusion}

The Agricultural Robotic System project is a significant advancement in agricultural technology. It aims to revolutionize traditional farming practices by minimizing manual labor, increasing efficiency, and reducing errors. The project integrates cutting-edge technologies such as the Yolov5 model for object detection and the CMUSphinx toolkit for speech recognition. The use of a pre-trained Chinese acoustic model has opened up new avenues for human-computer interaction, making the technology more accessible and user-friendly. Furthermore, the exploration of robot arm kinematics has provided valuable insights into the motion and operation of robotic arms, paving the way for more sophisticated and precise agricultural machinery.

The project has immense potential for future exploration. One area of interest could be the development of more advanced and diverse acoustic models, allowing for recognition of a wider range of languages and dialects. This would make the technology even more accessible to farmers around the world. Another promising direction is the further refinement of the object detection model. With advancements in AI and machine learning, it may be possible to train the model on a larger and more diverse dataset, enabling it to recognize a wider variety of crops and perform more complex tasks. Moreover, the study of robot arm kinematics could be extended to include more complex movements and operations, potentially leading to the development of robotic arms capable of performing tasks such as pruning or pest control.

In essence, the Agricultural Robotic System project stands as a beacon of innovation in agricultural technology. Its success thus far is a testament to the transformative power of robotics and AI in agriculture. As the project continues to evolve and overcome its challenges, it is set to play a pivotal role in shaping the future of farming practices worldwide. The journey of exploration and discovery is just beginning, and the potential for growth and advancement is immense.
%
%
%
%





\bibliographystyle{splncs04}
\bibliography{Bibliography}
\end{document}